\documentclass[11pt,a4paper]{article}
\usepackage[hyperref]{acl}
\usepackage{times}
\usepackage{latexsym}
\usepackage{multirow}
\usepackage{booktabs}
\usepackage{graphicx}
\usepackage{xcolor}
\usepackage{enumitem}
\usepackage{hyperref}
\usepackage{amssymb}
\usepackage{inconsolata}
\usepackage{pifont}
\newcommand{\xmark}{\ding{55}}

\usepackage{microtype}

\def\aclpaperid{1816} 

\newcommand{\ignore}[1]{}

\title{Label Semantic Aware Pre-training for Few-shot Text Classification}

\author{Aaron Mueller$^{1}$\Thanks{ Work done as an intern at Amazon Web Services.} \hspace{0.75cm}
  Jason Krone$^2$ \hspace{0.75cm}
  Salvatore Romeo$^2$ \\
  \bf{Saab Mansour$^2$ \hspace{0.75cm}
  Elman Mansimov$^2$ \hspace{0.75cm}
  Yi Zhang$^2$ \hspace{0.75cm}
  Dan Roth$^{3,2}$} \\
  $^1$Department of Computer Science, Johns Hopkins University \\
  $^2$Amazon Web Services AI Labs \\
  $^3$Department of Computer \& Information Science, University of Pennsylvania \\
  \texttt{amueller@jhu.edu}, \texttt{\{kronej,romeos,saabm,mansimov,yizhngn,drot\}@amazon.com}
  }

\date{}

\begin{document}
\maketitle
\begin{abstract}
In text classification tasks, useful information is encoded in the label names.
Label semantic aware systems have leveraged this information for improved text classification performance during fine-tuning and prediction. However, use of label-semantics during pre-training has not been extensively explored. 
We therefore propose \textbf{L}abel \textbf{S}emantic \textbf{A}ware \textbf{P}re-training (LSAP) to improve the generalization and data efficiency of text classification systems. LSAP incorporates label semantics into pre-trained generative models (T5 in our case) by performing secondary pre-training on labeled sentences from a variety of domains. As domain-general pre-training requires large amounts of data, we develop a filtering and labeling pipeline to automatically create sentence-label pairs from unlabeled text. We perform experiments on intent (ATIS, Snips, TOPv2) and topic classification (AG News, Yahoo!\ Answers). LSAP obtains significant accuracy improvements over state-of-the-art models for few-shot text classification while maintaining performance comparable to state of the art in high-resource settings.
\end{abstract}

\section{Introduction}

Large pre-trained language models have enabled better performance on many NLP tasks---especially in few-shot settings \citep{brown2020language,schick2021exploiting,wu-dredze-2020-languages}. More informative representations of textual inputs often leads to much higher downstream performance on NLP applications, which explains the rapid and general adoption of models such as (Ro)BERT(a) \citep{devlin2019bert,liu2019roberta}, GPT-2 \citep{radford2019language}, and T5 \citep{t5}. However, while these models are often used to effectively encode inputs, fewer works have attempted to give models access to informative representations of \emph{labels} as well.

\begin{figure}
    \centering
    \includegraphics[width=\linewidth]{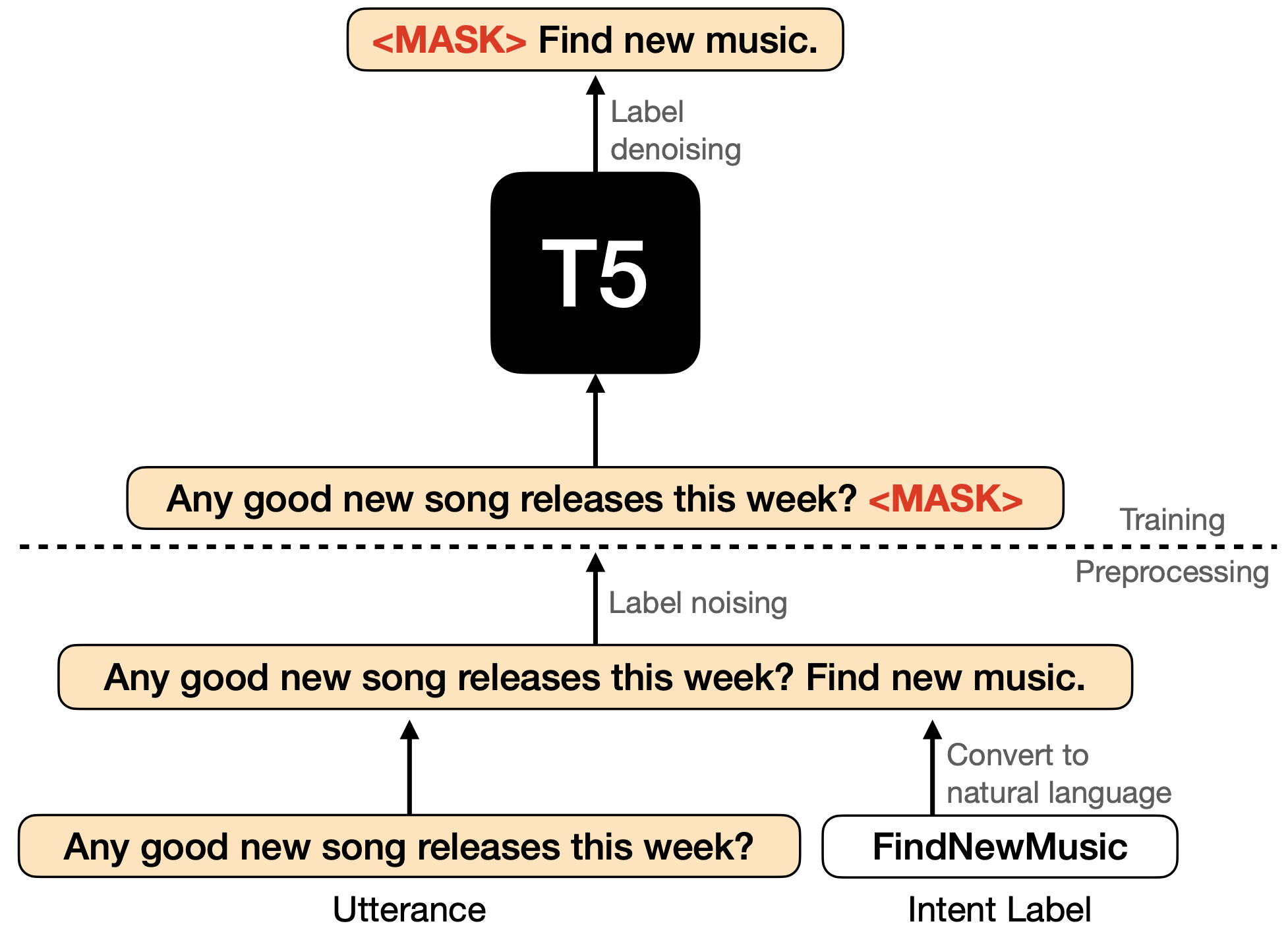}
    \caption{Overview of our approach, label semantic aware pre-training (LSAP). We collect utterance-intent pairs and create new pairs from unlabeled Reddit and Twitter data, convert the intents to natural language, concatenate the utterance and intent, noise the concatenated sequence, and train a sequence-to-sequence model to denoise the sequence.}
    \label{fig:approach_overview}
\end{figure}

Most discriminative approaches to text classification only give the model access to label indices. A recent stream of work has obtained significant improvements in structured prediction tasks by using sequence-to-sequence (seq2seq) models to generate labels \citep{athiwaratkun2020gsl, paolini2021tanl}. Yet these generative approaches make use of label semantics---the meaning of class label names---only during fine-tuning and prediction. Thus, we propose \textbf{L}abel \textbf{S}emantic \textbf{A}ware \textbf{P}re-training (LSAP)  to incorporate label semantics as well as input-label associations into the pre-training step (Figure~\ref{fig:approach_overview}). Our experiments show that LSAP yields higher performance with fewer fine-tuning examples in a variety of domains.
 
Our contributions include the following:
\begin{enumerate}[noitemsep]
    \item A method to incorporate label semantics into generative models during pre-training. 
    \item A method for creating utterance-intent pairs for label semantic aware pre-training from unlabeled noisy data.
    \item State-of-the-art few-shot performance on intent and topic classification datasets.
\end{enumerate}

Our code is publicly available.\footnote{\url{https://github.com/amazon-research/label-aware-pretrain}}

\section{Related Work}

\textbf{Label semantics} has been leveraged in many settings and tasks to improve performance and robustness, even before dense embedding representations became standard: \citet{chang2008labsem} achieve over 80\% accuracy on binary text classification tasks without any labeled training examples by giving naïve Bayes classifiers rich semantic representations \citep{gabrilovich2007esa} of labels. \citet{song2014dataless} use a similar approach for hierarchical and multinomial classification, finding that dataless procedures approached the performance of (and sometimes outperformed) supervised approaches.

More recently, label semantics based on dense embedding representations have become popular---especially with the rise of contextualized word embeddings \citep{peters2018elmo}. One stream of work integrates label representations using label vectors and label attention: label-wise attention networks (LWAN; \citealp{mullenbach2018explainable}) are designed for datasets with very large structured label spaces, and where multiple labels can apply to a potentially very long document. \citet{mullenbach2018explainable} generate label feature vectors and use an LWAN based on convolutional neural networks (CNNs); \citet{rios2018lwan} extend the attention mechanism for zero-shot settings. \citet{chalkidis2020bertlwan} use BERT to embed the labels for an LWAN.\footnote{Note that BERT-Base often performs similarly to or better than BERT-LWAN, and XLNet performs significantly better than BERT-Base in text classification settings that more closely resemble our setting \citep{yang2020xlnet}. We therefore opt to compare to XLNet.} In contrast, our label space is flatter and much smaller, and the input texts are much shorter on average.

Another label-semantic-aware approach for massively multi-label text classification is CLESS \citep{rethmeier2022dataefficient}. CLESS performs contrastive pre-training from scratch on a question-answering dataset with a large and sparse label space. Using contrasts between positive and negative answer embeddings, they obtain performance comparable to or better than RoBERTa using only in-domain data. However, CLESS pre-trains and classifies on the same data, and is meant for classification in one domain with a very large label space; our setting contains a larger mismatch between pre-training (where we have domain-general data and a large label space) and fine-tuning (where we have domain-specific data and a small label space). We want our approach to be more domain-general and to leverage the simplicity and effectiveness of token and span reconstruction objectives---for example, masked language modeling \citep{devlin2019bert} and span denoising \citep{t5}. Thus, we opt to perform secondary pre-training on data from a variety of domains using an existing model.

Other work integrates label embeddings into short-text intent and topic classification systems, more similarly to our task. \citet{gaonkar2020labsem} use label embeddings from BERT and a label attention mechanism to improve emotion classification accuracy; \citet{ma2022labsemner} also employ label embeddings for improved few-shot named entity recognition. Generative approaches like those of \citet{rongali2020generate,athiwaratkun2020gsl,paolini2021tanl} implicitly make use of label semantics for text and token classification tasks by generating the labels at prediction time.
\citet{rastogi2019scalable} use embeddings of human-defined schema which guide a dialogue state tracking system. 

\textbf{Few-shot text classification} entails performing classification after training or tuning a model on only a few examples from the training split of an evaluation set. Recent approaches to this include (Ro)BERT(a)-based \citep{topv2} and especially XLNet-based \citep{yang2020xlnet} classifiers, prototypical networks \citep{snell2017prototypical}, dynamic memory induction networks \citep{geng2020dynamic}, and generative classification (discussion follows).

Text classification contains more specific subtasks. Here, we evaluate on topic classification (TC; labeling text with its general domain, e.g.\ ``world news'') and intent classification (IC; labeling intentful text with what it is trying to accomplish, e.g.\ ``book flight''). The scarcity of IC data in many domains prevents the use of many neural text classification methods \citep{krone2020fewexamples}. In IC, the input is a conversational utterance and the output is a label describing what the user intends to do; for example, given a set of intents \{\texttt{BookHotel}, \texttt{BookFlight}, \texttt{CheckAccount}\} and an input utterance ``I would like a flight to NYC next month'', the model should classify the utterance as \texttt{BookFlight}. Recent approaches to few-shot IC include dual encoders \citep{cer2018use,henderson2020convert,casanueva2020efficient}, combining prototypical networks with meta-learning \citep{krone2020fewexamples}, a nearest-neighbor discriminative method \citep{zhang2020discriminative}, and span-level contextualized embedding retrieval \citep{yu2021fewshot}.

\textbf{Generative text classification} has become more feasible with the advent of large pre-trained language models \citep{radford2019language,brown2020language}. Here, one tunes a model to generate a natural language label given an input sequence, which reduces the train-test gap and does not require architectural changes.

Pattern-exploiting training (PET; \citealp{schick2021exploiting,schick2021small}) entails formatting train and test examples as cloze-style prompts, where the label is generally one word. Here, a language model can see the inputs and the embeddings of the output classes during tuning and prediction, though unlike our approach, there is no domain-general training step. Prompt tuning \citep{lester2021power,gao2021lmbff} improves PET by automatically optimizing the prompt design. Our approach is even more widely applicable in that we give the model concatenated utterances and labels and allow it to transduce to variable-length label sequences, rather than reformatting the evaluation data into a cloze format where the label must be one word/token.

Sequence-to-sequence (seq2seq) approaches based on pointer/copy mechanisms \citep{rongali2020generate} or T5 \citep{t5} tend to be effective and more data-efficient for text and token classification in semantic parsing tasks, including slot labeling \citep{athiwaratkun2020gsl} and entity-relation extraction \citep{paolini2021tanl}. In this method, one trains or fine-tunes a seq2seq model to transduce from an unlabeled text sequence to a sequence with labeled spans (or to just the label). Our approach is based on T5, but unlike prior work, we perform a \emph{label semantic aware} secondary pre-training step on a variety of datasets before fine-tuning.

\section{Approach}
Our approach, LSAP, performs a secondary pre-training step with T5 on a large scale collection of (pseudo-)labeled examples; see Figure~\ref{fig:approach_overview} for an overview. In addition to training on existing labeled datasets (\S\ref{sec:data}), we (1) filter unlabeled data using a \textbf{dialogue act classifier} (\S\ref{sec:wild_data}); (2) pseudo-label the utterances that pass the filter using an \textbf{intent generator} (\S\ref{sec:wild_data}); and (3) perform secondary pre-training on the labeled and pseudo-labeled data using T5 (\S\ref{sec:pretrain_method}). \S\ref{sec:data} details how we collect, select, and label pre-training data. \S\ref{sec:pretrain_method} describes the pre-training formats we test in our experiments.

\subsection{Data}\label{sec:data}
Our pre-training data (Table~\ref{tab:data_pretrain}) consists of utterances with intent labels. We use intentful utterances for pre-training because intent labels are often more specific, informative, and varied than sentiment or topic labels. For example, consider the intents \texttt{BookFlight} and \texttt{ViewAirfare} versus the topic \texttt{AirTravel}.
Our pre-training corpus combines gold (human-labeled), silver (heuristically labeled), and bronze (pseudo-labeled) data.

We first collect gold datasets: here, we use a set of non-public benchmark datasets
as well as PolyAI Banking \citep{coope2020polyai}, which yields approximately 130K training examples containing over 1,200 unique intents.\footnote{We ensure that there is no overlap between the gold data and our evaluation sets.}

To supplement this data, we add the silver WikiHow intents \citep{zhang2020wikihow} dataset, which is heuristically labeled. Each WikiHow example consists of an utterance, the longest step in a WikiHow article, and an intent label (the article title with ``How To'' removed). Training on this dataset has been shown to improve few-shot IC performance in a variety of domains \citep{zhang2020wikihow}.

\subsubsection{Pseudo-labeling Noisy Data}\label{sec:wild_data}
\begin{figure}[!ht]
    \centering
    \includegraphics[width=0.6\linewidth]{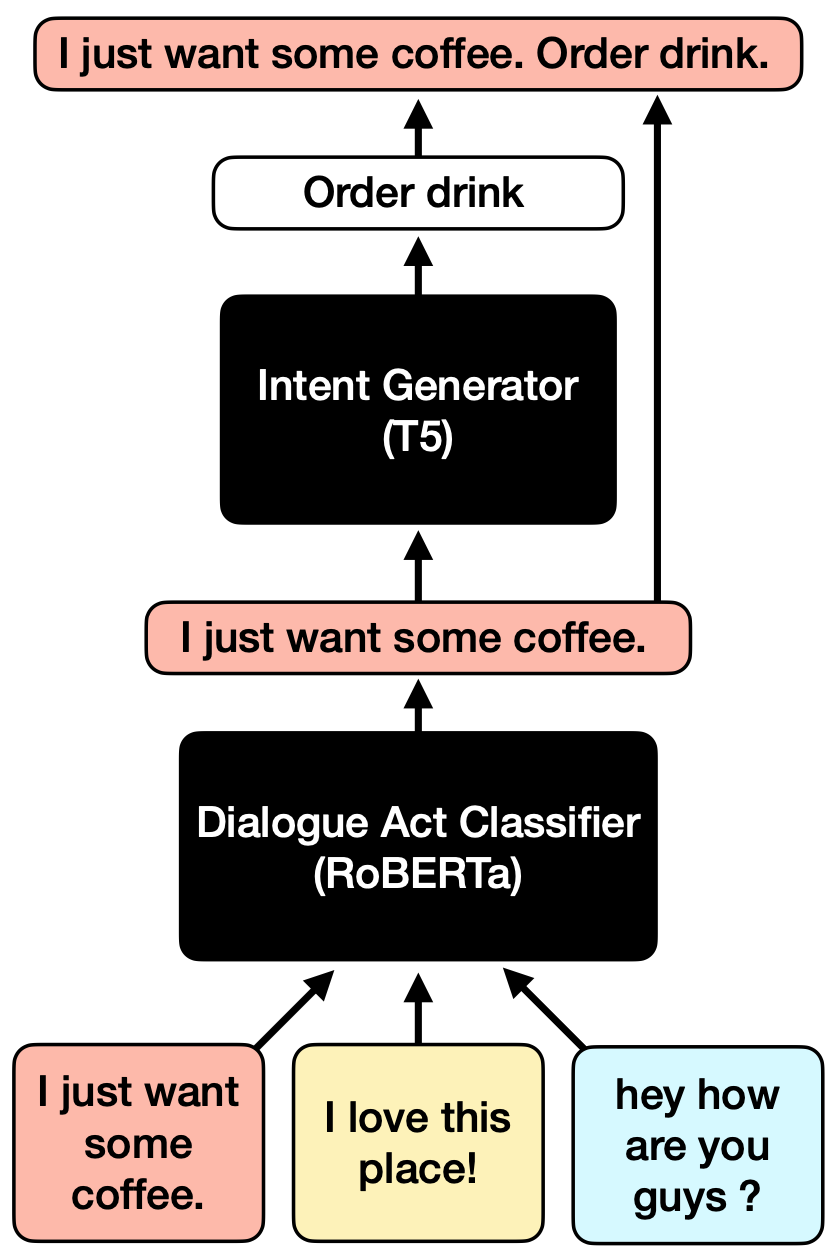}
    \caption{Our pipeline for creating utterance-intent pairs from unlabeled conversational data. We automatically filter data for intentful utterances using a dialogue act classifier, and then run filtered utterances through a T5-based intent generator. Each filtered utterance and its respective intent is concatenated to create a pre-training example.}
    \label{fig:wild_data}
\end{figure}

Gold and silver conversational datasets are scarce and tend to focus on one or a few narrow domains. To obtain more data from a larger variety of domains, we propose a filtering and labeling pipeline for converting unlabeled conversational data into pseudo-labeled ``bronze'' pre-training data. See Figure~\ref{fig:wild_data} for an overview.

\begin{table*}
    \centering
    \resizebox{0.99\linewidth}{!}{
    \begin{tabular}{llp{6cm}rrr}
    \toprule
    \textbf{Dataset} & \textbf{Quality} & \textbf{Description} & \multicolumn{2}{r}{\textbf{Training Examples}} & \textbf{Intents} \\
    \midrule
    Internal benchmark data & Gold & Benchmark datasets. 15+ domains from English locales. & & 119,920 & 1,172 \\
    PolyAI Banking & Gold & Online banking queries. & & 10,016 & 77 \\
    \midrule
    WikiHow Intent Classification & Silver & Automatically labeled intent classification dataset. Intent is article title with "How to" removed, and utterance is the longest step in the article. & & 110,573 & 110,573 \\
    \midrule
    & & & \textbf{Pre-filter} & \textbf{Post-filter} \\\cmidrule{4-5}
    Customer Support on Twitter & Bronze & Conversations (tweets) between consumers and customer support agents on Twitter. & 2,811,774 & 446,309 & 122,909 \\
    Reddit PushShift & Bronze & Comments scraped from Reddit. & 100,000,000 & 680,000 & 220,786 \\
    \bottomrule
    \end{tabular}}
    \caption{Pre-training datasets. ``Quality'' refers to the source of the labels (human-labeled is gold, deterministically labeled is silver, probabilistically labeled is bronze). ``Pre-filter'' and ``Post-filter'' refer to the number of training examples before and after using the dialogue act classifier described in \S\ref{sec:wild_data}.}
    \label{tab:data_pretrain}
\end{table*}

We start by collecting conversational utterances. We first use the Customer Support on Twitter (CSTwitter) dataset;\footnote{\url{https://www.kaggle.com/thoughtvector/customer-support-on-twitter}} this consists of over 2.81M tweets to and from company customer support agents. We also use the Reddit PushShift dataset;\footnote{\url{https://files.pushshift.io/reddit/comments/}} this dataset is large, so we only download most recent comment dumps by date until reaching 100M comments.

\paragraph{Dialogue act classifier.} In a conversational dataset, many utterances will not be intentful, and thus will not lend themselves to informative labels. For example, the statement ``Hiking is an outdoor activity'' does not express a clear goal out-of-context (i.e., it is non-intentful), whereas ``Buy a plane ticket to NYC for next month'' is a command with a clear goal. Applying an intent label to a non-intentful utterance may lead to supervision that is harmful to downstream performance. To filter for intentful utterances, we tune a RoBERTa-based binary classifier (intentful vs.\ non-intentful)
using Multi-Domain Goal-Oriented Dialogue (MultiDoGO; \citealp{peskov2019multidogo}) and Schema-guided Dialogue (SGD; \citealp{rastogi2019scalable,kale2020template}). For MultiDoGO, we treat greetings/goodbyes, thank yous, and other generic intents as non-intentful/negative examples; we treat all other intents that are not out-of-domain as intentful/positive examples. For SGD, we treat any utterance tagged with INFORM 
intents as non-intentful; utterances with intents tagged as REQUEST are treated as intentful. When evaluating on a held-out set of MultiDoGO and SGD, the classifier achieves 98\% precision.

To evaluate the precision of the classifier on our newly filtered data, we randomly sample 150 
utterances (per dataset) tagged as intentful by the classifier and calculate the proportion that are actually intentful, as judged by human evaluation. For CSTwitter, we obtain 91\% precision; this high precision may be due to the dataset being composed primarily of intentful customer-service-focused queries. For Reddit, we initially obtained 54\% precision. We qualitatively find that the probability assigned by the classifier to the positive label correlates well with intentfulness, so for Reddit, we exclude all examples to which our classifier assigned a positive-label probability lower than the median for all utterances tagged as positive examples. After probability thresholding, we obtain 76\% precision on Reddit.

\paragraph{Intent generator.} To label the intentful utterances, we train a T5-based generative intent labeler. We fine-tune T5 on the gold and silver data to transduce from utterances to intents (e.g., ``intent classification: Find me a hotel in NYC'' $\rightarrow$ ``Book hotel''), and then apply this tuned model to the filtered utterances. We find that this model generates intents not seen in the training set for 37\% of the utterances. These novel intents often but do not always demonstrate lexical overlap with the utterance (e.g., ``i'm wondering if there's fireworks for sale there?'' $\rightarrow$ ``Fireworks for sale'') or add specific descriptors to intents from the training set (e.g., ``Find a job'' appears in the training set, but our model generates ``Find a job in the UK'' which does not).

\subsection{Pre-training Formats}\label{sec:pretrain_method}
\begin{figure}
    \centering
    \includegraphics[width=0.9\linewidth]{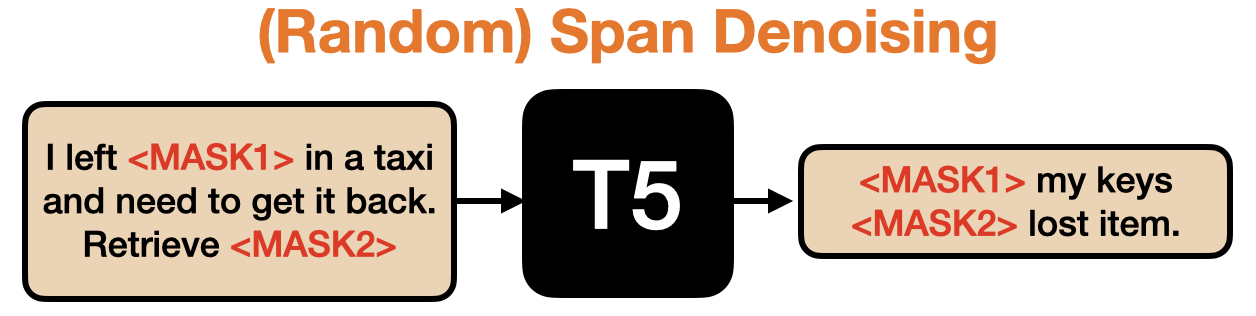}
    
    \vspace{2pt}
    
    \includegraphics[width=0.9\linewidth]{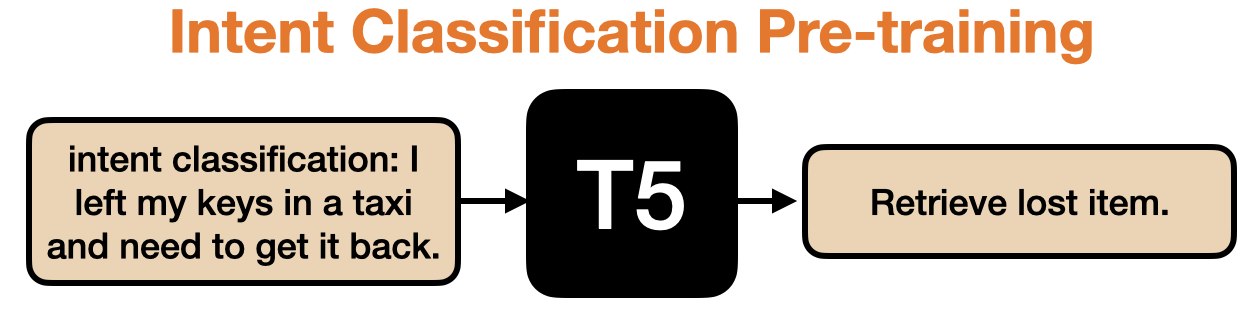}
    
    \vspace{1pt}
    
    \includegraphics[width=0.9\linewidth]{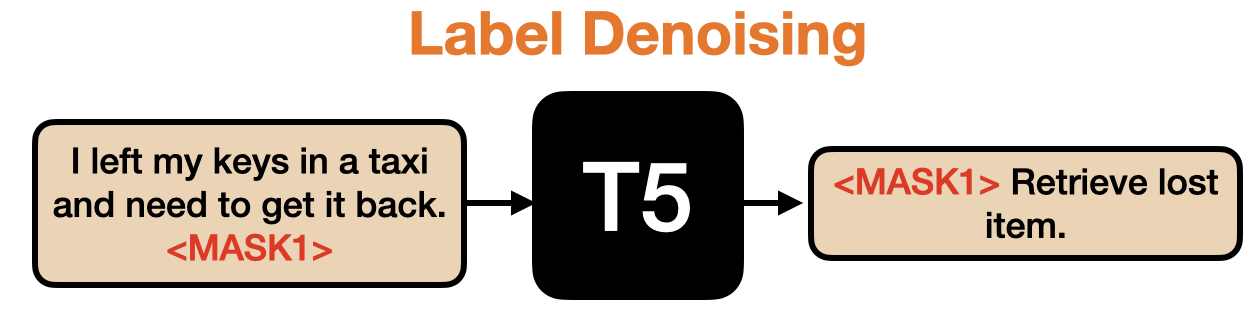}
    \caption{Pre-training formats that we test in our experiments. We find that label denoising is best.}
    \label{fig:pretrain_formats}
\end{figure}

We experiment with different pre-training formats, all but one of which are based on T5's span denoising objective. See Figure~\ref{fig:pretrain_formats} for an overview of our formats and an example of each.

First, we try \emph{random span denoising}. In this approach, we concatenate each intent label (in natural language format) to its associated utterance. Then, we randomly noise 15\% of the tokens\footnote{15\% is the proportion of tokens that are noised in the original pre-training setup \citep{t5}.} in the utterance-intent input sequence, reconstructing the contiguous noised spans in the output sequence. This is the same objective that T5 uses.\footnote{Unlike T5, we do not pack multiple sequences into single training examples. When we pack multiple utterance-intent pairs into training examples of length $\approx 512$, we find that performance drops sharply compared to training on individual utterance-intent pairs.}

Our second approach is \emph{intent classification (IC) pre-training}, where we supervise T5 with utterances and intents in the same format used for downstream supervised fine-tuning. Here, the input sequence is ``intent classification: '' followed by the utterance. The output sequence is the intent.

Finally, we implement \emph{label denoising}, where we train on utterances and their respective intents as before; however, instead of noising random spans, we deterministically noise the entire label sequence in the input sequence and reconstruct it in the output sequence. This is a way of framing the intent classification task as an unsupervised denoising task. This format is formally equivalent to IC pre-training in that we must transduce from utterances to intents, though we empirically show that the unsupervised denoising format matters a great deal for downstream IC performance.

\subsection{Evaluation Label Overlap is Rare}
We search for non-case-sensitive intent matches with Snips and ATIS in our pre-training dataset, finding that exact intent matches (or intents with Snips/ATIS intents as substrings) are very rare in our data: 682 examples (0.005\%) featured exact or substring overlap. We also search for the presence of \emph{any} lexical overlap between the intents in ATIS/Snips and the intents in our pre-training examples (i.e., if any word in an ATIS or Snips intent appears in a pre-training example's intent, we count it): less than 8,000 examples (0.6\%) in our pre-training dataset featured lexical overlap. Utterances tagged with exact intent matches were often not similar to the kinds of utterances seen in the evaluation sets; for example, \texttt{PlayMusic} is a label in Snips that typically refers to playing specific songs or artists, while the same label in our pre-training set often referred to requests to buy or play an instrument.

To check whether there was semantic overlap more broadly between the utterances in our pre-training set and those in Snips/ATIS, we obtain sentence embeddings for 5 randomly sampled utterances from each intent in Snips and ATIS, and for each utterance in our pre-training dataset. Sentence embeddings are obtained using a sentence-BERT model \citep{reimers2019sentence}; we use \texttt{all-MiniLM-L6-v2}, as it is both fast and performs best on semantic similarity tasks. We then calculate pairwise semantic similarity between the utterances in our pre-training set and the Snips/ATIS utterances by calculating the cosine similarity between sentence embedding pairs. When observing the most similar utterances, we do not often see much semantic overlap. For example, ``Play some Mf Doom from the sixties on pandora'' (in Snips) has the greatest sentence embeddings similarity to ``I love doom and won't let some launcher protest stop me from playing it'' (in our dataset); there is lexical overlap in the utterance, but the sentence in our dataset refers to playing a video game rather than a song. However, there were rare instances where semantic overlap was significant: for example, ``Add this tune to the Rock Save the Queen playlist'' (in Snips) was similar to ``Please add the Fresh Prince of Bel-Air song to this'' in our dataset.

\section{Experimental Setup}
We aim to understand whether LSAP can improve text classification performance over a variety of state-of-the-art (SOTA) baselines.

\begin{table}[h]
    \centering
    \resizebox{\linewidth}{!}{
    \begin{tabular}{lrrrc}
    \toprule
    Dataset & Train & Test & Classes & Balanced? \\
    \midrule
    Snips & 13,784 & 700 & 7 & \checkmark \\
    ATIS & 4,978 & 700 & 20 & \\
    TOPv2 (reminder) & 494 & 338 & 9 & \\
    TOPv2 (weather) & 177 & 148 & 4 & \\
    \midrule
    Yahoo!\ Answers & 1.4M & 60,000 & 10 & \checkmark \\
    AG News & 120,000 & 7,600 & 4 & \checkmark \\
    \bottomrule
    \end{tabular}}
    \caption{Evaluation datasets for intent classification (top) and topic classification (bottom). For TOPv2, we use the two provided low-resource domains as separate evaluation sets; we use the 25 Samples per Intent and Slot (SPIS) splits from \citet{topv2} as the maximum split size.}
    \label{tab:eval_data}
\end{table}

We evaluate LSAP on two text classification tasks: intent classification (IC) and topic classification (TC). IC is the most similar task to our pre-training objective. Our IC evaluation datasets include Snips \citep{coucke2018snips}, ATIS \citep{price1990atis}, and the low-resource \emph{reminder} and \emph{weather} domains provided in TOPv2 \citep{topv2}. Snips' intents are each from different domains. ATIS focuses on the flight domain; some intents appear only once or not at all in the training set, and some utterances are tagged with multiple intents (we count these as separate intents). For multi-class examples, we separate intents with the character ``\#'' (e.g., ``book flight \# airfare''). TOPv2 (\emph{reminder}) focuses on the creation and modification of personal reminders, while TOPv2 (\emph{weather}) focuses on queries regarding the weather and time of sunrise/sunset.

For TC, we evaluate on Yahoo!\ Answers (YA)\footnote{\url{https://www.kaggle.com/soumikrakshit/yahoo-answers-dataset}} and AG News.\footnote{\url{http://groups.di.unipi.it/~gulli/AG_corpus_of_news_articles.html}} YA consists of conversational user questions labeled with the general category of the question (e.g., ``Health'', ``Sports''). AG News consists of formal-register news articles labeled with their category (e.g., ``Business'', ``World'').

\begin{table*}[ht]
    \centering
    \resizebox{\linewidth}{!}{
    \begin{tabular}{lrrrrrrrrrrrr}
    \toprule
    & \multicolumn{3}{c}{Snips} & \multicolumn{3}{c}{ATIS} & \multicolumn{3}{c}{TOPv2 (reminder)} & \multicolumn{3}{c}{TOPv2 (weather)} \\\cmidrule(lr){2-4}\cmidrule(lr){5-7}\cmidrule(lr){8-10}\cmidrule(lr){11-13}
    \textbf{Format} & 1-shot & 4-shot & Full & 1-shot & 4-shot & Full & 1-shot & 4-shot & Full & 1-shot & 4-shot & Full \\
    \midrule
    Span denoising & 80.9 & 92.0 & \textbf{99.1} & 67.0 & 85.4 & \textbf{97.8} & 60.9 & 80.0 & \textbf{92.0} & 61.5 & 77.1 & 86.4 \\
    IC pre-training & 67.1 & 90.9 & 99.0 & 67.0 & 86.9 & 97.4 & 68.2 & 77.4 & 87.1 & 70.5 & 80.1 & 87.1 \\
    Label denoising & \textbf{88.7} & \textbf{93.5} & 99.0 & \textbf{68.7} & \textbf{87.8} & 97.6 & \textbf{69.0} & \textbf{80.6} & 91.4 & \textbf{72.7} & \textbf{81.4} & \textbf{89.1} \\
    \bottomrule
    \end{tabular}}
    \caption{1-shot, 4-shot, and full-resource accuracies across LSAP pre-training formats on each IC evaluation set. Label denoising is consistently best in few-shot settings while obtaining similar performance to other formats in full-resource settings. Subsequent tables refer to the label denoising variant as LSAP for brevity. Results for all models and split sizes in Appendix~\ref{app:ic_method}.}
    \label{tab:all_ic_method}
\end{table*}

\subsection{Fine-tuning}\label{sec:ft}
When fine-tuning our T5-based models, the input sequence consists of an ``intent classification:'' prefix followed by the utterance. The output sequence is the intent in a natural language format. For example: ``intent classification: Find me a flight from NYC to Baltimore.'' $\rightarrow$ ``Book flight''.\footnote{We also try label denoising fine-tuning to reduce the mismatch between pre-training and fine-tuning, but we find that this is not as effective as the traditional T5 fine-tuning setup. See Appendix~\ref{app:denoising_ft} for scores.}

We create few-shot splits of size $k$ for each evaluation set, where we sample $\leq k$ examples for each class $i$ from the set of labeled training examples $T$. More specifically, $\forall k \in \{1,2,4,8,16,32\}$,$\forall i\in T$, if $|T_i| \geq k$, we randomly sample $k$ training examples from $T_i$ \emph{without replacement}; if $|T_i| < k$, we simply use $T_i$. We ensure that smaller splits are subsets of larger splits such that we may perform more principled comparisons across split sizes.
In low-resource settings, the random seed can have a large impact on performance. We therefore average all accuracies over 5 random fine-tuning seeds.

See Appendix~\ref{app:hyperparams} for hyperparameters.

\subsection{Baselines}
XLNet \citep{yang2020xlnet} is an autoregressive Transformer-based \citep{vaswani2017attention} language model which sees various word order permutations of the inputs during pre-training. We append a linear layer to the mean-pooled output of XLNet to obtain an XLNet-based classifier. This model has achieved SOTA performance on many text classification datasets, including AG News.

SEQ2SEQ-PTR \citep{rongali2020generate} is based on sequence-to-sequence \citep{sutskever2014sequence} and pointer generator networks \citep{vinyals2015pointer,see2017pointer}; it achieves SOTA IC/SL performance on Snips, ATIS, and TOPv2. It transduces from an unlabeled utterance to a sequence with labeled spans (including the intent label, where the span is the entire sentence).

LM-BFF \citep{gao2021lmbff} is a prompt-tuning-based model for few-shot text classification. It uses automatically generated and optimized prompts (from T5) to perform generative classification (with RoBERTa) by predicting a masked token after the input. It achieves SOTA scores on few-shot binary/trinary sentiment classification and NLI tasks. We employ the best-performing \emph{prompt-tuning with demonstrations} approach, adapting it for IC by reducing the intent labels to single words.

T5 is the closest baseline to LSAP: our model differs only in the addition of the label aware pre-training step. We thus compare LSAP against T5 to understand the contribution of our label aware pre-training approach to downstream performance. T5 is comparable to TANL \citep{paolini2021tanl}, which is based on vanilla T5; TANL's SOTA performance on a variety of structured prediction tasks indicates that T5 as-is can be a strong baseline.

To test whether performance improvements with LSAP can be attributed to label semantics or domain adaptation on the utterances, we also present results for T5 adapted (with random span denoising) on only on the utterances and \emph{not} the intents in our pre-training data. This is equivalent to LSAP with random span denoising, but without the intent labels. We refer to this as ``T5 (adapted)''.

\section{Results}

Which pre-training format is best? We present 1-shot, 4-shot, and full-resource IC accuracies on each evaluation set in Table~\ref{tab:all_ic_method}. \textbf{Label denoising is the best-performing pre-training format in few-shot settings across all evaluation sets}. Differences in performance between LSAP formats decrease as we increase the fine-tuning set size; in full-resource settings, the difference between span denoising and label denoising is not significant. This suggests that explicitly demarcating intents from utterances during pre-training may help T5 better leverage the pre-training examples.\footnote{These results are stable across random samples of the few-shot splits. See Appendix~\ref{app:random_fewshot_samples}.} As ``label denoising'' performs best, we focus on that variant of LSAP from here on, though note that all LSAP formats outperform T5 and T5 (adapted).

Table~\ref{tab:snips_atis_ic_method} displays IC accuracies across models. Compared to T5, the T5 (adapted) baseline achieves consistently higher IC accuracies. However, T5 (adapted) is vastly outperformed by LSAP (by up to 18\% on Snips and 3\% on ATIS in 1-shot settings), even though the utterances are the same across these settings; this indicates that \textbf{much of the improvements may be attributed to label semantics}, and that domain adaptation is only responsible for a small portion of the improvement.

\begin{table}[ht]
\centering
\resizebox{\linewidth}{!}{
\begin{tabular}{lrrrrrrr}
    \toprule
    & \multicolumn{7}{c}{Snips: \textbf{Examples per Label}} \\\cmidrule{2-8}
    \textbf{Model} & 1 & 2 & 4 & 8 & 16 & 32 & Full \\
    \midrule
    XLNet & 70.0 & 77.6 & 88.1 & 92.9 & 96.2 & 96.9 & 99.0 \\
    LM-BFF & 75.3 & 82.2 & 88.3 & 94.0 & 96.5 & \textbf{97.6} & 98.9  \\
    SEQ2SEQ-PTR & 75.8 & 84.2 & 89.5 & 93.5 & 96.2 & 97.1 & 99.0 \\
    T5 & 71.1 & 79.5 & 89.5 & 92.9 & 95.2 & 96.5 & 99.0 \\
    T5 (adapted) & 74.9 & 81.3 & 91.2 & 94.4 & 96.2 & 96.9 & 98.9 \\
    LSAP & \textbf{88.7} & \textbf{90.5} & \textbf{93.5} & \textbf{94.8} & \textbf{96.7} & 97.3 & 99.0 \\
    \bottomrule
    \toprule
    & \multicolumn{7}{c}{ATIS: \textbf{Examples per Label}} \\\cmidrule{2-8}
    \textbf{Model} & 1 & 2 & 4 & 8 & 16 & 32 & Full \\
    \midrule
    XLNet & 24.1 & 46.8 & 70.2 & 77.2 & 92.4 & 94.4 & \bf{98.0} \\
    SEQ2SEQ-PTR & 15.6 & 31.6 & 45.8 & 77.0 & 83.3 & 95.3 & 97.4 \\
    T5 & 45.8 & 78.7 & 83.9 & 90.3 & 92.3 & 94.5 & 97.4 \\
    T5 (adapted) & 66.9 & 78.0 & 84.5 & 91.7 & 93.6 & 95.5 & 97.6 \\
    LSAP & \textbf{68.7} & \textbf{79.5} & \textbf{87.8} & \textbf{92.4} & \textbf{95.7} & \textbf{96.4} & 97.6 \\
    \bottomrule
\end{tabular}}
\caption{Mean intent classification accuracies across 5 seeds on Snips and ATIS at various few-shot split sizes. Smaller splits are subsets of larger splits. LSAP is consistently best in lower-resource settings while maintaining comparable performance to other models in higher-resource settings. Standard deviations and results for TOPv2 are in Appendix~\ref{app:ic_method}.}
\label{tab:snips_atis_ic_method}
\end{table}

\begin{table}[t]
\centering
\resizebox{\linewidth}{!}{
\begin{tabular}{lrrrrrrr}
    \toprule
    & \multicolumn{7}{c}{Yahoo!\ Answers: \textbf{Examples per Label}} \\\cmidrule{2-8}
    \textbf{Model} & 1 & 2 & 4 & 8 & 16 & 32 & Full \\
    \midrule
    XLNet & 23.9 & 42.4 & 44.0 & 52.8 & 62.3 & 65.9 & 77.6 \\
    T5 & 41.9 & 54.0 & 59.3 & 61.3 & 64.6 & 65.6 & \textbf{77.8} \\
    LSAP & \textbf{49.2} & \textbf{58.8} & \textbf{60.7} & \textbf{63.3} & \textbf{64.7} & \textbf{66.4} & 77.7 \\
    \bottomrule
    \toprule
    & \multicolumn{7}{c}{AG News: \textbf{Examples per Label}} \\\cmidrule{2-8}
    \textbf{Model} & 1 & 2 & 4 & 8 & 16 & 32 & Full \\
    \midrule
    XLNet & 55.3 & 63.5 & 69.7 & 79.4 & \textbf{85.4} & \textbf{86.8} & \textbf{95.6} \\
    T5 & 62.0 & 74.4 & 77.2 & 81.1 & 84.6 & 85.9 & 94.8 \\
    LSAP & \textbf{74.8} & \textbf{77.2} & \textbf{80.7} & \textbf{82.3} & \textbf{85.4} & 86.5 & 94.8 \\
    \bottomrule
\end{tabular}}
\caption{Few-shot text classification accuracies on Yahoo!\ Answers (top) and AG News (bottom). Smaller splits are subsets of larger splits. Our LSAP approach yields improvements over the SOTA XLNet model. LSAP also improves up over vanilla T5, indicating that our approach generalizes to topic classification.}
\label{tab:yahoo_agnews_acc_just_best}
\end{table}


\begin{table}
\centering
\resizebox{\linewidth}{!}{
\begin{tabular}{lrrrrrr}
    \toprule
    & \multicolumn{6}{c}{Snips: \textbf{Examples per Label}} \\\cmidrule{2-7}
    \textbf{Pre-training Data} & 1 & 2 & 4 & 8 & 16 & 32 \\
    \midrule
    None & 71.2 & 79.5 & 89.5 & 92.9 & 95.2 & 96.5 \\
    \midrule
    gold & 82.2 & 86.9 & 91.7 & 94.0 & 95.9 & 96.5 \\
    \hspace{2mm}+silver & 81.6 & 85.3 & 92.0 & 95.0 & 96.0 & 96.8 \\
    \hspace{4mm}+CSTwitter & 82.9 & 88.4 & 92.7 & 94.4 & \textbf{96.7} & 97.1 \\
    \hspace{6mm}+Reddit & \textbf{88.7} & \textbf{90.5} & 93.5 & 94.8 & \textbf{96.7} & 97.3 \\
    \midrule
    CSTwitter+Reddit & 80.9 & 90.1 & \textbf{93.7} & \textbf{95.2} & 96.6 & \textbf{97.6} \\
    C4 & 82.4 & 87.1 & 92.8 & 94.9 & 96.2 & 97.3 \\
    \bottomrule
\end{tabular}}
\caption{Pre-training dataset ablation using LSAP (with label denoising). We display intent classification accuracies on Snips. Each pre-training dataset improves performance, but we can recover most of the performance from our best dataset when using \emph{only} our bronze data.}
\label{tab:snips_ic_data}
\end{table}

We next observe that \textbf{generative approaches are generally more effective than discriminative approaches, especially in lower-resource settings.} The XLNet classifier does not have access to label semantics during fine-tuning, instead observing class indices and utterances only; all other approaches hence have access to more information during fine-tuning and prediction.

\textbf{LSAP induces better performance than LM-BFF and SEQ2SEQ-PTR} in very low-resource settings, though LM-BFF and SEQ2SEQ-PTR perform better than T5. Note that for ATIS, LM-BFF does not effectively handle the multi-class examples due to the single-word cloze format, nor the high number of similar labels due to the intended contrastive word use case; we thus do not compare to LM-BFF here.

\begin{table*}
\centering
\resizebox{0.9\linewidth}{!}{
\begin{tabular}{lccrrrrrr}
    \toprule
    & & & \multicolumn{6}{c}{Snips: \textbf{Examples per Label}} \\\cmidrule{4-9}
    \textbf{Model} & Shuffled pre-train labels? & Remapped eval labels? & 1 & 2 & 4 & 8 & 16 & 32 \\
    \midrule
    T5 & N/A & \checkmark & -18.0 & -9.9 & -0.9 & -0.4 & -1.4 & -0.2 \\
    T5 (adapted) & N/A & \checkmark & -21.6 & -11.9 & -7.8 & -2.9 & +0.1 & -0.1 \\
    LSAP & \xmark & \checkmark & -23.6 & -9.4 & -1.8 & -1.7 & 0.0 & -0.2 \\
    LSAP & \checkmark & \xmark & -42.4 & -27.0 & -15.7 & -14.2 & -17.4 & -14.1 \\
    LSAP & \checkmark & \checkmark & -40.6 & -31.9 & -17.9 & -10.6 & -15.4 & -8.9 \\
    \bottomrule
    \toprule
    & & & \multicolumn{6}{c}{ATIS: \textbf{Examples per Label}} \\\cmidrule{4-9}
    \textbf{Model} & Shuffled pre-train labels? & Remapped eval labels? & 1 & 2 & 4 & 8 & 16 & 32 \\
    \midrule
    T5 & N/A & \checkmark & -17.2 & -25.2 & -7.0 & -6.2 & -9.0 & -5.8 \\
    T5 (adapted) & N/A & \checkmark & -41.1 & -29.7 & -9.7 & -7.9 & -4.3 & -1.3 \\
    LSAP & \xmark & \checkmark & -35.7 & -29.6 & -10.6 & -6.9 & -4.5 & -2.8 \\
    LSAP & \checkmark & \xmark & -35.4 & -19.5 & -17.6 & -15.4 & -8.8 & -9.0 \\
    LSAP & \checkmark & \checkmark & -41.8 & -30.4 & -36.0 & -28.0 & -21.7 & -16.7 \\
    \bottomrule
\end{tabular} }
\caption{Absolute difference in intent classification accuracy relative to LSAP on Snips and ATIS at all few-shot split sizes after shuffling labels assigned to utterances in the pre-training set (random label semantics), remapping labels in the evaluation set (misleading label semantics), or both.}
\label{tab:snips_atis_misleading_labsem_diffs}
\end{table*}

As we increase the size of the fine-tuning splits, performance increases and converges across methods. Thus, the primary contribution of LSAP is inducing quicker generalization across domains.

Finally, we present topic classification accuracies for YA and AG News (Table~\ref{tab:yahoo_agnews_acc_just_best}). Our LSAP approach yields up to 35\% improvements on AG News over the SOTA XLNet model in 1-shot settings, and over 100\% on YA; we also maintain comparable performance to the SOTA XLNet baseline in full-resource settings. LSAP also improves up to 21\% on AG News and 18\% on YA over vanilla T5; this is evidence that \textbf{our pre-training procedure is not just tuning T5 to recognize utterance-intent associations: it is also teaching T5 how to be a better text classifier in general.} This also indicates that our procedure generalizes to labels that have not been seen during pre-training: topic labels are not present in the pre-training data, though some of these labels do appear as substrings of more specific intent labels (e.g., while ``Health'' does not appear as a label in our pre-training data, ``Get health information'' does appear).

Our model also generalizes well to joint IC/SL. See Appendix~\ref{app:joint_ic_sl}.

\subsection{Dataset Ablation}
Using LSAP with label denoising, we ablate over each pre-training dataset and observe the effect on downstream IC performance (Table~\ref{tab:snips_ic_data}).

The gold data improves few-shot IC performance over T5 by over 10\%. Adding the silver data does not significantly change performance. Adding CSTwitter improves performance in few-shot settings, though it also increases variance across random seeds. The best accuracies and lowest variances are achieved after adding the Reddit data.

Notably, we find that \textbf{pre-training on only automatically filtered and labeled examples still improves performance over T5.} To test whether this is due to CSTwitter and Reddit simply being well-suited to our evaluation sets, we try automatically filtering and labeling a less conversational dataset: Colossal Cleaned Common Crawl (C4; \citealp{t5}), the same dataset used for pre-training T5. We use the same number of training examples as in our combined CSTwitter+Reddit data for comparability (1,126,309 examples). We still observe performance improvements when pre-training on this dataset, and the difference between pre-training on this versus CSTwitter+Reddit is only significant at 2 examples per label. This is evidence that \textbf{our method for creating new pre-training examples is effective with different types of data, including non-conversational data}.
Note that adding C4 to the gold+silver+CSTwitter+Reddit data does not significantly improve performance; we therefore do not include it in our final pre-training dataset.

\subsection{The Importance of Meaningful Labels}\label{sec:misleading_labsem}

Including label sequences in the pre-training data results in large performance increases. Is this due solely to intent sequences being helpful independent information, or is the model learning useful associations between the inputs and labels? To investigate this, we experiment with misleading and random label semantics. For ``misleading label semantics,'' we remap the labels in an evaluation dataset by defining a bijective function from each intent to a randomly selected different intent, ensuring that identity mappings do not occur. In other words, if label $i$ is replaced with label $j$, all instances of $i$ are systematically converted to $j$ in the training \emph{and} test sets. We then fine-tune our models on these remapped-intent datasets and observe whether performance decreases, and by how much. Significant performance decreases indicate reliance on input-label associations.

We also define a ``random label semantics'' variant of LSAP, where we randomly shuffle the intents with respect to the utterances in the \emph{pre-training data} and then pre-train with label denoising on the shuffled data. Unlike the evaluation label remapping, this shuffling procedure preserves the number of instances of each intent such that our pre-training set is technically the same---the intents and utterances are simply mismatched.

Performance with misleading and/or random label semantics (Table~\ref{tab:snips_atis_misleading_labsem_diffs})\footnote{Misleading label semantics results for TOPv2 are in Appendix~\ref{app:misleading_labsem}.}  decreases considerably across datasets, and this is especially apparent in the low-resource case. Label denoising seems the most sensitive to utterance-label associations. Performance drops decrease with increasing few-shot split sizes. This suggests that \textbf{our models do rely on utterance-intent associations to achieve high IC performance, and that these associations are more important in lower-resource settings than higher-resource settings.}

\section{Conclusions}
We have proposed a pre-training approach for leveraging the semantic information inherent in labels. Our method improves few-shot text classification performance across domains while maintaining high performance in full-resource settings. This approach is fairly general and could potentially be extended to more structured semantic parsing tasks by annotating some of the pre-training examples, or perhaps by simply including diverse labeled examples from a variety of tasks in the pre-training step. Future work could investigate extending this method to pre-training from scratch, as well as tuning the utterance and label formats. One could also use demonstrations to achieve better performance with fewer gradient updates.

\section*{Acknowledgments}
We thank Giovanni Paolini, Ben Athiwaratkun, and the reviewers for their thoughtful feedback on earlier versions of this work.

\bibliographystyle{acl_natbib}
\bibliography{anthology,acl2021}

\clearpage

\appendix

\newpage

\section{Joint Intent Classification and Slot Labeling}\label{app:joint_ic_sl}
Intent classification (IC) and slot labeling (SL) are often performed simultaneously. For sequence-to-sequence models, this can be done by transducing from an unlabeled utterance to a labeled utterance, as in \citet{athiwaratkun2020gsl}. To evaluate whether our model can be used as an effective IC/SL system, we use the same approach, substituting our model for TANL. This tuning setup is more distinct from our pre-training task than the IC setup, so we hypothesize that performance gains over TANL will be slightly smaller here on IC than in the approach in the main paper. We also hypothesize that we will see similar SL scores as for TANL.

Our results (Table~\ref{tab:joint_icsl}) indicate that in this setting, our pre-training approach still results in increased intent classification accuracy over TANL. As found before, the performance improvements are most noticeable in the few-shot setting. We also find that slot labeling performance is mostly maintained after pre-training, including in the 1-shot setting. Future work could investigate ways to integrate slot labeling supervision into pre-training for improving performance on both IC and SL.

\begin{table}[h]
    \centering
    \resizebox{\linewidth}{!}{
    \begin{tabular}{clrrrr}
    \toprule
    & & \multicolumn{2}{c}{\bf{Intent Class.}} & \multicolumn{2}{c}{\bf{Slot Labeling}} \\\cmidrule(lr){3-4}\cmidrule(lr){5-6}
    & \bf{Model} & \bf{ATIS} & \bf{Snips} & \bf{ATIS} & \bf{Snips} \\
    \midrule
    \multirow{4}{*}{\bf{Full}} & Joint BERT & 98.6 & 97.5 & \bf{97.0} & \bf{96.1} \\
    & ELMO+BiLSTM & \bf{99.3} & 97.4 & 93.9 & 95.6 \\
    & TANL & 99.0 & 97.0 & 96.9 & \bf{96.1} \\
    & LSAP (label denoising) & 99.1 & \bf{97.6} & 96.8 & \bf{96.1} \\
    \midrule
    \multirow{2}{*}{\bf{1-shot}} & TANL & 78.8 & 88.5 & \bf{36.0} & \bf{81.7} \\
    & LSAP (label denoising) & \bf{82.3} & \bf{89.3} & 35.8 & 80.8 \\
    \bottomrule
    \end{tabular}}
    \caption{Intent classification accuracy and slot labeling F1 on ATIS and Snips.}
    \label{tab:joint_icsl}
\end{table}

\section{Stability Across Random Samples of the Few-shot Splits}\label{app:random_fewshot_samples}
\begin{figure}[ht]
    \centering
    \includegraphics[width=0.9\linewidth]{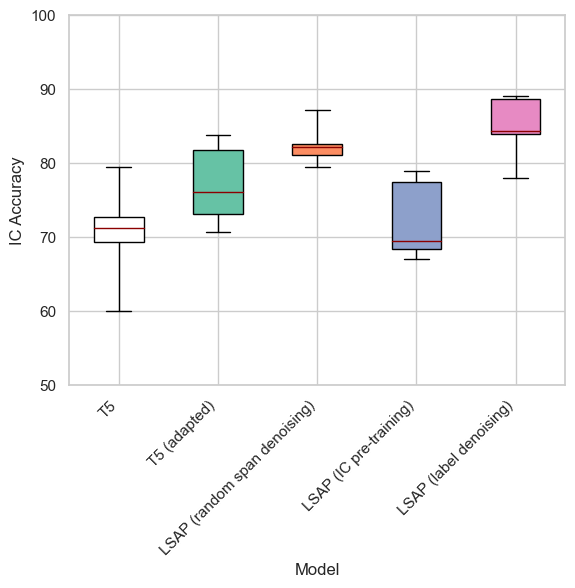}
    \centering
    \includegraphics[width=0.9\linewidth]{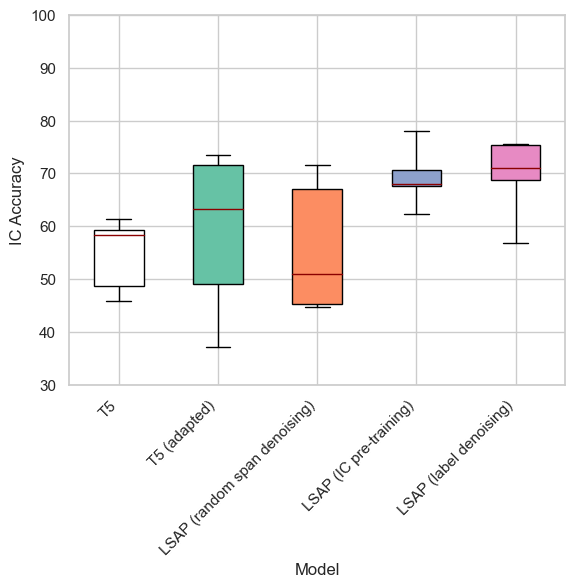}
    \caption{Intent classification accuracy distribution across 5 random samples of the 1-shot fine-tuning set for Snips (top) and ATIS (bottom).}
    \label{fig:snips_atis_samples}
\end{figure}
In few-shot settings, the selection of fine-tuning examples can make a large difference in downstream performance. Here, we present IC accuracy distributions across 5 random samples of the 1-shot results (Figure~\ref{fig:snips_atis_samples}) to understand whether the relative performance of each approach is stable given different fine-tuning set samples. Specifically, we present macroaverages across random fine-tuning set samples after averaging performance across random seeds; this is so that we average over more stable estimates of performance within each fine-tuning set.

Performance on Snips across random 1-shot fine-tuning samples is largely stable. All of our original conclusions regarding which models are better than others still hold: label denoising is best among our label semantic aware methods, and each label semantic aware approach (except IC pre-training) beats the non-label semantic aware baselines. Performance on ATIS is more variable across random 1-shot samples, but the relative performance of models is still largely the same: IC pre-training is more competitive on this evaluation set and random span denoising does not beat the T5 (adapted) baseline, but label denoising is still best.

\section{Pre-training Method Comparison: Full Results}\label{app:ic_method}
Here, we present intent classification accuracies for ATIS (Table~\ref{tab:atis_ic_method}) the low-resource domains of TOPv2 (Tables~\ref{tab:top_reminder_ic_method},\ref{tab:top_weather_ic_method}) using each of our pre-training approaches. Our results here indicate that the relative performance of each method is stable across evaluation sets: \textbf{label denoising is consistently the best approach in the lowest-resource setting.} Intent classification pre-training is often second-best, while random span denoising is consistently least effective among our pre-training formats (but still improves performance over vanilla T5 and the T5 (adapted) baseline for \emph{all} evaluation datasets).

\begin{table*}[ht]
\centering
\resizebox{\linewidth}{!}{
\begin{tabular}{lrrrrrrr}
    \toprule
    & \multicolumn{7}{c}{Snips: \textbf{Examples per Label}} \\\cmidrule{2-8}
    \textbf{Model} & 1 & 2 & 4 & 8 & 16 & 32 & Full \\
    \midrule
    XLNet & 70.0 & 77.6 & 88.1 & 92.9 & 96.2 & 96.9 & 99.0 \\
    LM-BFF & 75.3 (3.0) & 82.2 (3.6) & 88.3 (3.4) & 94.0 (0.5) & 96.5 (0.5) & \textbf{97.6} (0.4) & 98.9  \\
    SEQ2SEQ-PTR & 75.8 (2.1) & 84.2 (2.1) & 89.5 (0.9) & 93.5 (1.0) & 96.2 (0.3) & 97.1 (0.4) & 99.0 \\
    T5 & 71.1 (2.2) & 79.5 (2.8) & 89.5 (1.1) & 92.9 (1.7) & 95.2 (0.7) & 96.5 (0.6) & 99.0 \\
    T5 (adapted) & 74.9 (4.7) & 81.3 (3.1) & 91.2 (1.0) & 94.4 (0.5) & 96.2 (0.3) & 96.9 (0.3) & 98.9 \\
    \midrule
    \multicolumn{7}{l}{LSAP} \\
    \hspace{8pt} Span denoising & 80.9 (2.0) & 85.7 (1.6) & 92.0 (1.3) & 94.3 (1.2) & 96.5 (0.2) & 97.1 (0.7) & \textbf{99.1} \\
    \hspace{8pt} IC pre-training & 67.1 (2.1) & 83.2 (1.4) & 90.9 (0.8) & 94.5 (0.5) & 96.5 (0.4) & 97.4 (0.3) & 99.0 \\
    \hspace{8pt} Label denoising & \textbf{88.7} (1.5) & \textbf{90.5} (0.9) & \textbf{93.5} (0.8) & \textbf{94.8} (0.7) & \textbf{96.7} (0.4) & 97.3 (0.3) & 99.0 \\
    \bottomrule
\end{tabular}}
\caption{Pre-training method comparison. Mean intent classification accuracies across 5 seeds on Snips at various few-shot split sizes. Smaller splits are subsets of larger splits. LSAP is consistently best in lower-resource settings while maintaining comparable performance to other models in higher-resource settings.}
\label{tab:snips_ic_method_app}
\end{table*}

\begin{table*}
\centering
\resizebox{\linewidth}{!}{
\begin{tabular}{lrrrrrrr}
    \toprule
    & \multicolumn{7}{c}{ATIS: \textbf{Examples per Label}} \\\cmidrule{2-8}
    \textbf{Model} & 1 & 2 & 4 & 8 & 16 & 32 & Full \\
    \midrule
    XLNet & 24.1 & 46.8 & 70.2 & 77.2 & 92.4 & 94.4 & \bf{98.0} \\
    SEQ2SEQ-PTR & 15.6 (5.7) & 31.6 (4.0) & 45.8 (7.1) & 77.0 (5.2) & 83.3 (2.1) & 95.3 (0.7) & 97.4 \\
    T5 & 45.8 (18.2) & 78.7 (7.1) & 83.9 (2.9) & 90.3 (1.3) & 92.3 (1.7) & 94.5 (1.0) & 97.4 \\
    T5 (adapted) & 66.9 (4.4) & 78.0 (6.1) & 84.5 (2.6) & 91.7 (1.0) & 93.6 (1.6) & 95.5 (0.8) & 97.6 \\
    \midrule
    \multicolumn{4}{l}{LSAP} \\

    \hspace{8pt} Span denoising & 67.0 (6.0) & 77.6 (6.2) & 85.4 (3.1) & 91.4 (1.4) & 94.6 (1.3) & 95.5 (0.9) & 97.8 \\
    \hspace{8pt} IC pre-training & 67.0 (3.0) & 77.0 (1.1) & 86.9 (2.1) & 90.9 (1.0) & 94.9 (0.8) & 95.5 (0.8) & 97.4 \\
    \hspace{8pt} Label denoising & \textbf{68.7} (14.3) & \textbf{79.5} (8.1) & \textbf{87.8} (2.5) & \textbf{92.4} (7.9) & \textbf{95.7} (0.7) & \textbf{96.4} (0.8) & 97.6 \\
    \bottomrule
\end{tabular}}
\caption{Pre-training method comparison. Intent classification accuracy on ATIS at various few-shot split sizes. Smaller splits are subsets of larger splits. Each score is averaged over 5 fine-tuning runs. Note that the high standard deviation for T5 (label denoising) in the 1-shot setting is because the accuracies skew high; three seeds yield accuracies over 80\%, one at 76.9\%, and one at 60\%.}
\label{tab:atis_ic_method}
\end{table*}

\begin{table*}
\centering
\resizebox{0.8\linewidth}{!}{
\begin{tabular}{lrrrrr}
    \toprule
    & \multicolumn{5}{c}{TOPv2 (reminder): \textbf{Examples per Label}} \\\cmidrule{2-6}
    \textbf{Model} & 1 & 2 & 4 & 8 & 25SPIS \\
    \midrule
    XLNet & 33.5 & 39.8 & 51.9 & 76.0 & 89.0 \\
    T5 & 51.9 (5.3) & 61.6 (6.3) & 72.3 (2.5) & 83.3 (3.0) & 91.7 \\
    T5 (adapted) & 58.5 (3.6) & 66.6 (2.3) & 73.4 (4.0) & 85.2 (2.6) & 90.8 \\
    \midrule
    \multicolumn{4}{l}{LSAP} \\

    \hspace{8pt} Span denoising & 60.9 (5.0) & 68.4 (4.0) & 80.0 (1.6) & \textbf{88.6} (1.6) & 92.0 \\
    \hspace{8pt} IC pre-training & 68.2 (3.1) & 67.8 (1.8) & 77.4 (1.5) & 86.2 (1.3) & 87.1 \\
    \hspace{8pt} Label denoising & \textbf{69.0} (4.2) & \textbf{71.3} (3.8) & \textbf{80.6} (2.6) & 87.7 (0.5) & 91.4 \\
    \bottomrule
\end{tabular}}
\caption{Pre-training method comparison. Intent classification accuracy on TOPv2 (\emph{reminder} domain) at various few-shot split sizes. Smaller splits are subsets of larger splits. Each score is averaged over 5 fine-tuning runs.}
\label{tab:top_reminder_ic_method}
\end{table*}

\begin{table*}
\centering
\resizebox{0.8\linewidth}{!}{
\begin{tabular}{lrrrrr}
    \toprule
    & \multicolumn{4}{c}{TOPv2 (weather): \textbf{Examples per Label}} \\\cmidrule{2-6}
    \textbf{Model} & 1 & 2 & 4 & 8 & 25SPIS \\
    \midrule
    XLNet & 44.9 & 54.4 & 68.7 & 79.7 & 87.8 \\
    T5 & 53.2 (6.1) & 66.5 (11.8) & 74.4 (5.8) & 83.1 (0.5) & 87.1 \\
    T5 (adapted) & 61.2 (3.8) & 72.0 (4.5) & 77.4 (1.8) & 84.9 (1.3) & 83.7 \\
    \midrule
    \multicolumn{4}{l}{LSAP} \\

    \hspace{8pt} Span denoising & 61.5 (8.1) & 73.2 (3.0) & 77.1 (3.3) & 83.7 (0.7) & 86.4 \\
    \hspace{8pt} IC pre-training & 70.5 (3.6) & \textbf{77.4} (1.5) & 80.1 (1.0) & \textbf{83.8} (0.5) & 87.1 \\
    \hspace{8pt} Label denoising & \textbf{72.7} (1.3) & \textbf{77.4} (2.0) & \textbf{81.4} (1.1) & 83.4 (0.8) & \textbf{89.1} \\
    \bottomrule
\end{tabular}}
\caption{Pre-training method comparison. Intent classification accuracy on TOPv2 (\emph{weather} domain) at various few-shot split sizes. Smaller splits are subsets of larger splits. Each score is averaged over 5 fine-tuning runs.}
\label{tab:top_weather_ic_method}
\end{table*}

\section{The Importance of Meaningful Labels: Full Results}\label{app:misleading_labsem}
Here, we present intent classification performance with misleading label semantics (after the random label remapping procedure described in \S\ref{sec:misleading_labsem}) for
the low-resource domains of TOPv2 (Tables~\ref{tab:reminder_misleading_labsem_diffs},\ref{tab:weather_misleading_labsem_diffs}). We present accuracy \emph{differences} after remapping labels in order to observe how performance changes when using intent names that are not semantically related to the utterances they classify.

We find that our results are mostly consistent across evaluation sets: our best models rely more on utterance-intent associations and are thus more sensitive to label remapping, as evidenced by higher performance drops after label remapping. However, there appears to be variance across datasets with respect to reliance on intent labels: the performance drops for TOPv2 in both domains are much larger than for ATIS and Snips, providing evidence that \textbf{utterance-label associative information is much more important for some datasets than others}. We also find larger performance drops for our model pre-trained with the IC pre-training format than for our best model pre-trained with label denoising. Thus, \textbf{the best model is not necessarily the most reliant on utterance-label associations}.

The ``random label semantics'' baselines are sensitive to misleading label semantics for TOPv2 (reminder) and ATIS, as indicated by consistently large accuracy drops after label remapping. This is not the case for TOPv2 (weather) nor Snips in low-resource settings, where performance differences tend to be positive or closer to 0. The performance drop is much smaller for this baseline than for other models (indicating lower associative sensitivity), but it is still notable that the model was still able to rely on associations between utterances and intents after shuffling; perhaps this is due to the similarity of each intent label in these highly domain-specific datasets, though this would not explain the lack of sensitivity to utterance-label associations in TOPv2 (weather) since this is also highly domain-specific. Future work could more specifically investigate the source of these label sensitivity differences across datasets.

\begin{table*}
\centering
\resizebox{0.8\linewidth}{!}{
\begin{tabular}{lccrrrr}
    \toprule
    & & & \multicolumn{4}{c}{\small{TOPv2 (reminder): \textbf{Examples per Label}}} \\\cmidrule{4-7}
    \textbf{Model} & Shuffled pre-train labels? & Remapped eval labels? & 1 & 2 & 4 & 8 \\
    \midrule
    T5 & N/A & \checkmark & -13.3 & -16.1 & -25.8 & -21.6 \\
    T5 (adapted) & N/A & \checkmark & -25.2 & -26.8 & -26.4 & -17.2 \\
    LSAP & \xmark & \checkmark & -38.6 & -34.7 & -28.2 & -16.9 \\
    LSAP & \checkmark & \xmark & -44.5 & -29.9 & -30.2 & -20.0 \\
    LSAP & \checkmark & \checkmark & -53.4 & -40.3 & -46.5 & -23.2 \\
    \bottomrule
\end{tabular}}
\caption{Absolute difference in intent classification accuracy relative to LSAP on TOPv2 (\emph{reminder} domain) at all few-shot split sizes after shuffling labels assigned to utterances in the pre-training set, remapping labels in the evaluation set, or both.}
\label{tab:reminder_misleading_labsem_diffs}
\end{table*}

\begin{table*}
\centering
\resizebox{0.8\linewidth}{!}{
\begin{tabular}{lccrrrr}
    \toprule
    & & & \multicolumn{4}{c}{\small{TOPv2 (weather): \textbf{Examples per Label}}} \\\cmidrule{4-7}
    \textbf{Model} & Shuffled pre-train labels? & Remapped eval labels? & 1 & 2 & 4 & 8 \\
    \midrule
    T5 & N/A & \checkmark & -7.2 & -20.9 & -1.4 & -10.1 \\
    T5 (adapted) & N/A & \checkmark & -19.5 & -28.7 & -17.0 & -13.6 \\
    LSAP & \xmark & \checkmark & -34.3 & -37.3 & -19.7 & -8.7 \\
    LSAP & \checkmark & \xmark & -43.4 & -37.1 & -17.2 & -17.0 \\
    LSAP & \checkmark & \checkmark & -37.1 & -34.7 & -22.7 & -18.4 \\
    \bottomrule
\end{tabular}}
\caption{Absolute difference in intent classification accuracy relative to LSAP on TOPv2 (\emph{weather} domain) at all few-shot split sizes after shuffling labels assigned to utterances in the pre-training set, remapping labels in the evaluation set, or both.}
\label{tab:weather_misleading_labsem_diffs}
\end{table*}

\section{Error Analysis}
\begin{figure}
    \centering
    \includegraphics[width=\linewidth]{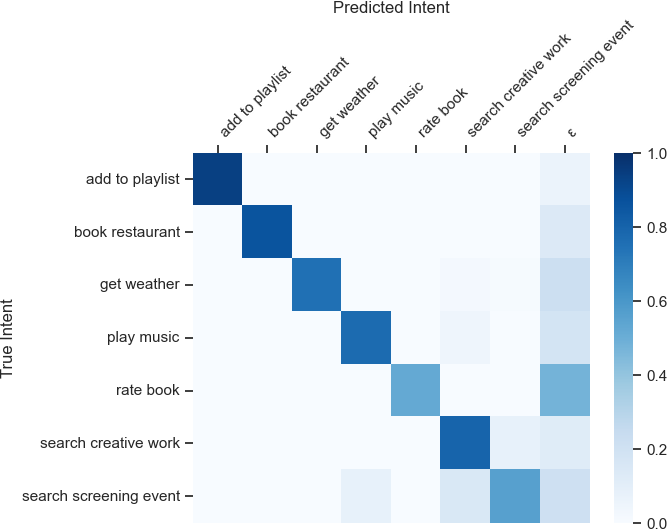}

    \includegraphics[width=\linewidth]{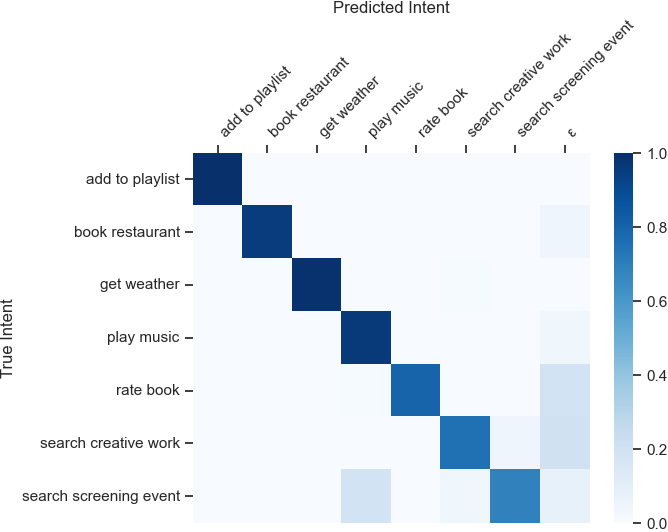}
    \caption{Confusion matrices displaying 1-shot results on Snips using vanilla T5 (top) and LSAP with label denoising (bottom). The fine-tuning setups are identical. ``$\epsilon$'' refers to generated labels outside the set of labels in Snips.}
    \label{fig:conf_mats}
\end{figure}

What kinds of errors does T5 make before and after pre-training? We present confusion matrices before and after our best label semantic aware pre-training approach (Figure~\ref{fig:conf_mats}). When using the same fine-tuning setup and hyperparameters, vanilla T5 tends to generate more out-of-domain intent labels than our best model. These out-of-domain generations are typically mergers between two intents: Snips contains \texttt{PlayMusic} and \texttt{RateBook} intents, and vanilla T5 often generates \texttt{RateMusic}. After pre-training, this is much less frequent.

Vanilla T5 also seems to assign higher prior probabilities to specific intents, despite seeing class-balanced tuning data; for example, \texttt{SearchCreativeWork} is generated for utterances from a variety of intents, as indicated by the slight vertical stripe in Figure~\ref{fig:conf_mats}. After pre-training, this problem is almost non-existent.

\section{Label Denoising Fine-tuning}\label{app:denoising_ft}
Here, we compare the performance of LSAP (label denoising pre-training) when fine-tuned using the traditional T5 format (task prefix, no masking) and when using the label denoising format (concatenate the document and its label in the source sequence, mask the label sequence, and reconstruct the label in the output sequence). See Tables~\ref{tab:denoising_ft} and \ref{tab:denoising_ft_top}.

\begin{table}[ht]
\centering
\resizebox{\linewidth}{!}{
\begin{tabular}{lrrrrrr}
    \toprule
    & \multicolumn{6}{c}{Snips: \textbf{Examples per Label}} \\\cmidrule{2-7}
    \textbf{Model} & 1 & 2 & 4 & 8 & 16 & 32 \\
    \midrule
    LSAP & & & & & & \\
    \hspace{8pt} FT & \textbf{88.7} & \textbf{90.5} & \textbf{93.5} & \textbf{94.8} & 96.7 & 97.3 \\
    \hspace{8pt} LDFT & 75.2 & 84.9 & 92.0 & 93.7 & \textbf{96.9} & \textbf{97.6} \\
    \bottomrule
    \toprule
    & \multicolumn{6}{c}{ATIS: \textbf{Examples per Label}} \\\cmidrule{2-7}
    LSAP & & & & & & \\
    \hspace{8pt} FT & \textbf{68.7} & \textbf{79.5} & \textbf{87.8} & 92.4 & \textbf{95.7} & \textbf{96.4} \\
    \hspace{8pt} LDFT & 59.7 & 76.3 & 85.7 & \textbf{93.0} & 95.1 & 95.8 \\
    \bottomrule
    \toprule
    & \multicolumn{6}{c}{Yahoo!\ Answers: \textbf{Examples per Label}} \\\cmidrule{2-7}
    \textbf{Model} & 1 & 2 & 4 & 8 & 16 & 32\\
    \midrule
    LSAP & & & & & & \\
    \hspace{8pt} FT & \textbf{49.2} & \textbf{58.8} & \textbf{60.7} & \textbf{63.3} & 64.7 & \textbf{66.4} \\
    \hspace{8pt} LDGT & 39.4 & 54.1 & 58.4 & 62.2 & \textbf{65.3} & \textbf{66.4} \\
    \bottomrule
    \toprule
    & \multicolumn{6}{c}{AG News: \textbf{Examples per Label}} \\\cmidrule{2-7}
    \textbf{Model} & 1 & 2 & 4 & 8 & 16 & 32 \\
    \midrule
    LSAP & & & & & & \\
    \hspace{8pt} FT & \textbf{74.8} & \textbf{77.2} & \textbf{80.7} & \textbf{82.3} & \textbf{85.4} & \textbf{86.5} \\
    \hspace{8pt} LDFT & 68.1 & 72.0 & 76.0 & 81.1 & 84.0 & 86.0 \\
    \bottomrule
    \end{tabular}}
\caption{Mean intent classification accuracies for normal fine-tuning (FT; as in \S\ref{sec:ft}) and label denoising fine-tuning (LDFT) across 5 seeds on Snips, ATIS, Yahoo!\ Answers, and AGNews at various few-shot split sizes. Smaller splits are subsets of larger splits.}
\label{tab:denoising_ft}
\end{table}

\begin{table}
\centering
\resizebox{\linewidth}{!}{
\begin{tabular}{lrrrrr}
    \toprule
    & \multicolumn{4}{c}{\small TOPv2 (reminder): \textbf{Ex.\ per Label}} \\\cmidrule{2-5}
    \textbf{Model} & 1 & 2 & 4 & 8 \\
    \midrule
    LSAP & & & & \\
    \hspace{8pt} FT & \textbf{69.0} & \textbf{71.3} & \textbf{80.6} & 87.7 \\
    \hspace{8pt} LDFT & 65.8 & 69.3 & 74.6 & 86.2 \\
    \bottomrule
    \toprule
    & \multicolumn{4}{c}{\small TOPv2 (weather): \textbf{Ex.\ per Label}} \\\cmidrule{2-5}
    \textbf{Model} & 1 & 2 & 4 & 8 \\
    \midrule
    LSAP & & & & \\
    \hspace{8pt} FT & \textbf{72.7} & 77.4 & \textbf{81.4} & 83.4 &\\
    \hspace{8pt} LDFT & 70.6 & \textbf{78.4} & 80.8 & \textbf{83.5} \\
    \bottomrule
\end{tabular}}
\caption{Mean intent classification accuracies for normal fine-tuning (FT; as in \S\ref{sec:ft}) and label denoising fine-tuning (LDFT) across 5 seeds on TOPv2 (\emph{reminder} and \emph{weather} domains) at various few-shot split sizes. Smaller splits are subsets of larger splits.}
\label{tab:denoising_ft_top}
\end{table}

In lower-resource settings, regular fine-tuning significantly outperforms label denoising fine-tuning across evaluation sets; an exception is TOPv2 (\emph{weather}) at 2 examples per label, but the difference in performance is not large here. In higher-resource settings, label denoising fine-tuning begins to achieve comparable performance to regular fine-tuning (and sometimes outperforms regular fine-tuning). Nonetheless, accuracy differences in higher-resource settings are not large, and regular fine-tuning performs significantly better on average. We therefore opt to use regular fine-tuning when comparing to baselines.

\section{Hyperparameters}\label{app:hyperparams}
Our experiments are based on the huggingface implementation \citep{wolf2020transformers} of T5.

For secondary pre-training, we use initial learning rate $5\times 10^{-4}$ and batch size $128$. We tune over the number of training epochs $\in [1,8]$, finding $3$ epochs to generally be best.

During fine-tuning, we use init.\ LR $5\times 10^{-4}$ and batch size $1$.\footnote{We use a small batch size due to the small size of the $1$-shot splits. We do not observe significant performance differences when using batch size 2 or 4.} We tune over the number of fine-tuning epochs $\in [1,16]$ for the largest few-shot split, typically finding $2$ epochs to be best. Once we have the best setting for the largest split, we double the number of tuning epochs for each halving of the split size such that the number of tuning steps is similar for all split sizes. All other hyperparameters are huggingface defaults.

\end{document}